\PassOptionsToPackage{table}{xcolor}
\documentclass[sigconf,screen,nonacm]{acmart}
\setcopyright{none}
\AtBeginDocument{%
  }
\usepackage{float}

\begin{document}

\title[EC-RAG for Long Video Understanding]{EC-RAG: Event Chain Retrieval-Augmented Generation for Long Video Understanding}

\author{Yuhao Qin}
\email{qinyuhaoswu@163.com}
\affiliation{%
  \institution{Northwestern Polytechnical University}
  \city{Xi'an}
  \country{China}
}

\author{Junbo Wang}
\email{jbwang@nwpu.edu.cn}
\affiliation{%
  \institution{Northwestern Polytechnical University}
  \city{Xi'an}
  \country{China}
}

\author{Yuke Li}
\email{liyuke@nwpu.edu.cn}
\affiliation{%
  \institution{Northwestern Polytechnical University}
  \city{Xi'an}
  \country{China}
}

\author{Yining Zhu}
\email{yiningzhu@nwpu.edu.cn}
\affiliation{%
  \institution{Northwestern Polytechnical University}
  \city{Xi'an}
  \country{China}
}

\renewcommand{\shortauthors}{Qin et al.}

\begin{abstract}
 Current large video-language models (LVLMs) still face challenges when dealing with long videos, mainly because frames are often processed independently, making it difficult to capture temporal dependencies across events. Although retrieval-augmented approaches have been introduced to provide additional context, most of them operate at the frame or snippet level, which limits their ability to model how events evolve over time and relate to each other. In this paper, we propose Event Chain Retrieval-Augmented Generation (EC-RAG), a training-free framework that organizes video content into an explicit event chain before question answering. Instead of retrieving isolated frames or text segments, EC-RAG first partitions the video into semantically coherent segments, represents each segment using multi-modal signals, and then links them into a structured chain that preserves temporal order and captures inter-event relationships. Given a query, the system identifies relevant events within this chain and gathers supporting evidence from the associated modalities. Our approach offers several practical advantages: (i) event-level abstraction that better reflects how video content is naturally structured, enabling more reliable localization compared to frame-level retrieval; (ii) structured multi-modal fusion that aggregates speech, text, and visual cues at the event level, allowing complementary information to be more effectively utilized during reasoning; and (iii) plug-and-play compatibility with existing LVLM backbones, requiring no additional training or reliance on proprietary models. Experiments on Video-MME, MLVU, and LongVideoBench show that this event-centric design consistently outperforms frame-level retrieval baselines, highlighting the importance of modeling temporal structure for long-video understanding.
\end{abstract}


\maketitle

\section{Introduction}
With the rapid advancement of Large Video-Language Models (LVLMs), significant progress has been made in video understanding tasks~\cite{liu2023visual, chen2024sharegpt4v, liu2026kangaroo, zhang2023video, zhang2024llavahddivinghighresolutionlarge, lin2024vila}. However, when processing long videos spanning tens of minutes to hours, current LVLMs face critical challenges: the limited context window restricts the number of frames that can be processed, while simply increasing sampled frames leads to information redundancy and reasoning degradation.

Recent studies have sought to extend the context length of LVLMs for long video understanding, such as LongLLaVA~\cite{long-llava-qwen2-7b} scales token capacity to ingest more frames. However, this brute-force expansion proves brittle—Video-MME~\cite{fu2025video} benchmark shows accuracy drops when additional frames are supplied beyond a threshold. This outcome suggests that simply increasing the number of sampled frames not only leads to information redundancy but also imposes additional challenges for the model to handle complex reasoning.
\begin{figure}[tbp]
  \centering
  \includegraphics[width=\linewidth]{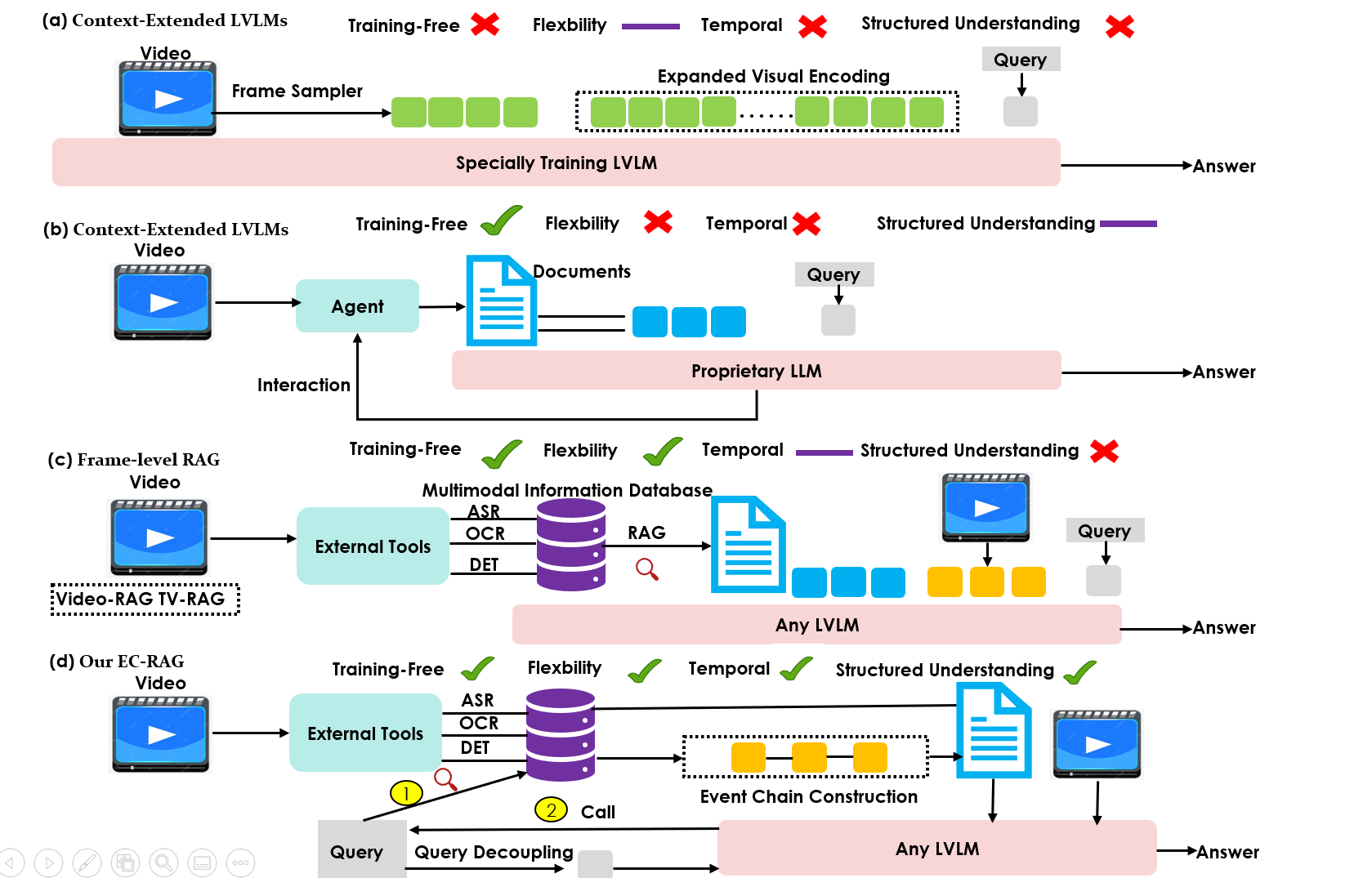} 
  \caption{Advantages of our EC-RAG. EC-RAG provides an event-centric, training-free pipeline with structured temporal understanding that is easily compatible with any LVLM.}
  \Description{}
\end{figure}
A parallel research direction employs Retrieval-Augmented Generation (RAG) to supplement video queries with externally retrieved information. Early efforts in this direction~\cite{liu2024llavanext, zhang2024llavanextvideo} demonstrate that augmenting visual inputs with auxiliary textual information can effectively enhance video understanding, particularly for content that is difficult to capture through visual frames alone. These methods typically extract three types of auxiliary texts from videos: automatic speech recognition (ASR) for transcribing spoken dialogue and narration, optical character recognition (OCR) for capturing on-screen text such as titles and labels, and object detection (DET) for identifying visual entities and their spatial arrangements. Building upon this foundation, Video-RAG~\cite{luo2024video} proposes visually-aligned auxiliary texts from OCR, ASR, and DET, achieving notable improvements with minimal computational overhead. TV-RAG~\cite{cao2025tv} further introduces temporal-window retrieval with entropy-based keyframe selection to handle semantic drift in long videos. More recently, AdaVideoRAG~\cite{xue2025adavideorag} employs adaptive difficulty routing and graph databases for complex multi-hop reasoning. Despite these advances, existing video RAG methods share a fundamental limitation: they retrieve at the frame level or text segment level, treating video as a collection of independent units rather than a coherent narrative of interconnected events. Long videos inherently contain sequences of causally connected events—a lecture progresses through topics, a movie unfolds through scenes, a cooking video follows recipe steps. Without modeling this event structure, current methods produce fragmented retrieval results that fail to support questions requiring temporal reasoning.

Based on this observation, we focus on the following question: \textit{Can we design a video RAG framework that explicitly captures temporal event dynamics and inter-event relationships for coherent long video understanding?}

To this end, we propose Event Chain Retrieval-Augmented Generation (EC-RAG), a training-free framework that organizes video content into explicit event chains before question answering. Instead of retrieving isolated frames or text snippets, EC-RAG first partitions the video into semantically coherent temporal segments, aggregates multi-modal signals (ASR, OCR, DET) within each segment to form events, and links them into a structured chain that preserves temporal order. For multi-modal extraction, we introduce CLIP-Variance guided keyframe selection that combines query-semantic relevance with visual informativeness, ensuring selected frames are both relevant and information-rich. Given a user query, EC-RAG performs event-level localization to identify the most relevant events along with their temporal context, and gathers supporting evidence from associated modalities for answer generation.

We evaluate EC-RAG on Video-MME~\cite{fu2025video} , MLVU~\cite{zhou2024mlvu} , and LongVideoBench~\cite{wu2024longvideobench} benchmarks. By applying EC-RAG to five different open-source LVLMs, we achieve an average performance improvement of 4\% on Video-MME. Furthermore, compared with existing frame-level retrieval methods such as Video-RAG and TV-RAG, EC-RAG consistently demonstrates superior performance, validating the effectiveness of our event-chain-based approach. Our contributions are summarized as follows:
\vspace{-4pt}
\begin{itemize}
\item We propose EC-RAG, a novel framework that explicitly models temporal event chains for long video understanding. Unlike existing methods operating at frame or snippet level, EC-RAG captures inter-event relationships and enables coherent reasoning over extended temporal spans.
\item We introduce CLIP-Variance guided keyframe selection that jointly considers semantic relevance and visual informativeness, improving the quality of multi-modal information extraction for event construction.
\item We design a query-guided event localization mechanism that identifies relevant events within the temporal chain and aggregates multi-modal evidence, providing structured context for LVLM answer generation.
\end{itemize}
\vspace{-4pt}

\section{Related Work}

\subsection{Video Understanding with Large Language Models}
The emergence of powerful LLMs has spurred efforts to develop comprehensive video-language systems. Early works like Video-ChatGPT~\cite{maaz2024video} and VideoChat~\cite{li2025videochat} adopt frame-wise feature extraction followed by temporal aggregation. Video-LLaVA~\cite{lin2024video} introduces a unified projection to bridge visual and textual modalities, while LLaVA-NeXT-Video~\cite{zhang2024llavanextvideo} further refines this through video-specific adaptation. These models perform well on short videos but struggle with longer content due to frame sampling constraints.

To address this limitation, recent methods explore expanding the context window for extended sequences. LongVA~\cite{zhang2024long} and Long-LLaVA~\cite{long-llava-qwen2-7b} leverage pre-training on lengthy documents to develop transferable sequence modeling capabilities. INTP~\cite{shang2024interpolating} reorganizes visual tokens to accommodate more information within fixed context limits. However, densely sampling frames quickly becomes computationally prohibitive, and the high redundancy in video data often limits the benefits of processing more frames.

\subsection{LLM-based Agents for Video Understanding}
An alternative paradigm employs LLMs as controllers to orchestrate specialized tools for video comprehension~\cite{gupta2023visual, min2024morevqa, suris2023vipergpt, wang2024videoagent, zhang2024simple}. Representative methods such as VideoAgent~\cite{fan2024videoagent} and MM-VID~\cite{lin2023mm} enable dynamic interaction with video content through iterative retrieval and reasoning. While flexible, these approaches introduce substantial latency from multiple tool calls and often rely on proprietary APIs, limiting broader adoption in open-source settings.

\subsection{Retrieval-Augmented Generation for Video}
RAG-based methods retrieve relevant information from pre-constructed databases to augment video understanding. Video-RAG~\cite{luo2024video} extracts multimodal cues including speech transcripts and detected objects to build searchable knowledge bases. TV-RAG~\cite{cao2025tv} enhances this with temporal-aware retrieval, while AdaVideoRAG~\cite{xue2025adavideorag} employs adaptive routing for queries of varying complexity. Current approaches, however, retrieve content at the frame or segment granularity without capturing the temporal progression of events, limiting performance on questions requiring understanding of how actions unfold across time.
\begin{figure*}[t]
  \centering
  \includegraphics[width=\linewidth]{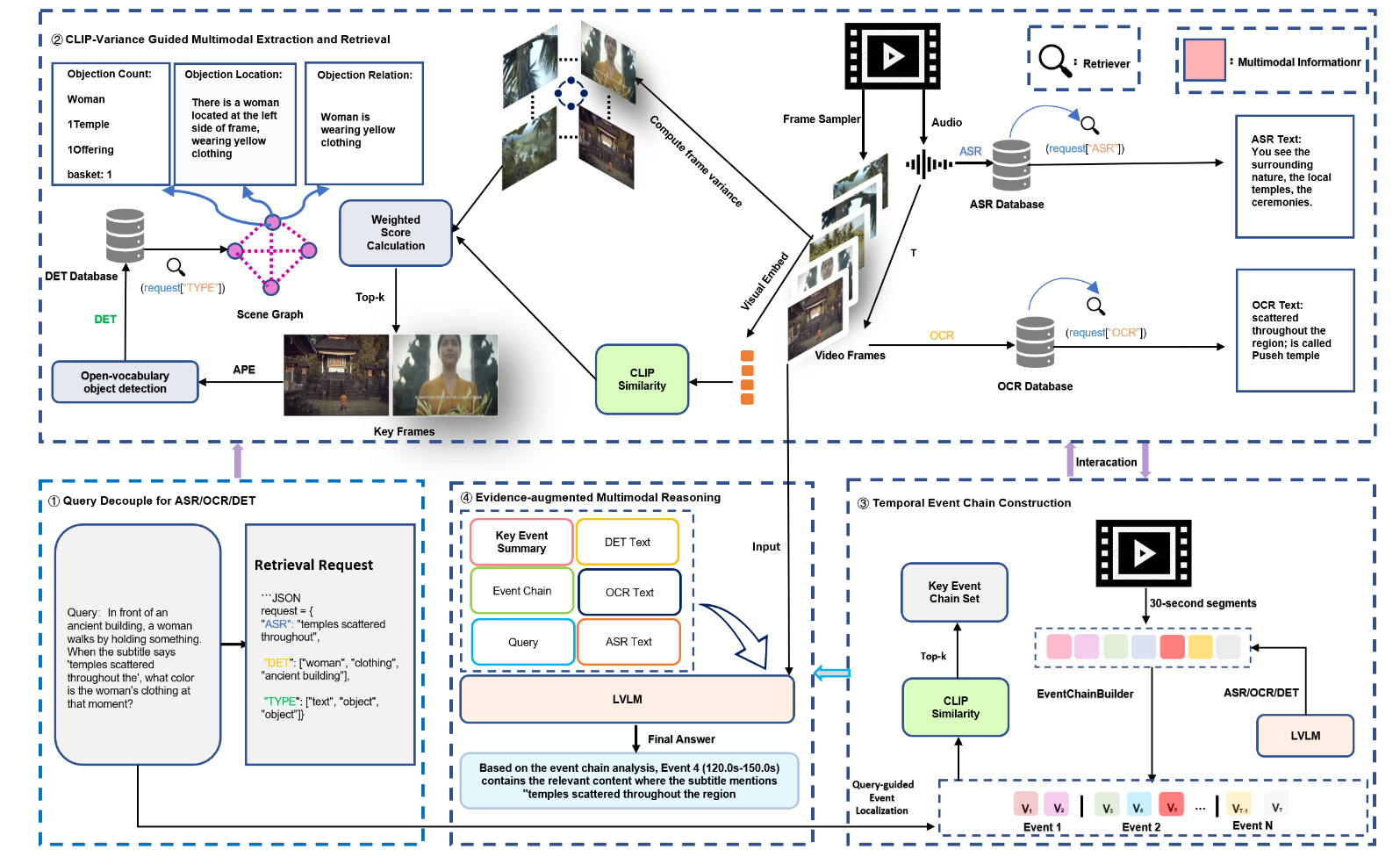} 
  \caption{Overview of the EC-RAG framework. The pipeline comprises four stages: (i) Query Decoupling parses the user question into modality-specific retrieval requests for ASR, OCR, and DET; (ii) CLIP-Variance Guided Multimodal Extraction selects informative keyframes through weighted scoring of visual similarity and frame variance, then extracts speech transcripts, on-screen text, and scene graphs via open-vocabulary object detection; (iii) Temporal Event Chain Construction segments the video into 30-second units and builds a chain of semantically coherent events, where each event integrates its associated multimodal evidence with explicit temporal relations; (iv) Evidence-augmented Multimodal Reasoning performs query-guided event localization to retrieve relevant events, then feeds the event chain summary, located events, and raw evidence to the LVLM for answer generation. By organizing video content as structured event sequences, EC-RAG captures temporal dynamics and inter-event dependencies for complex video understanding.}
  \Description{}
  \label{fig:framework}
\end{figure*}

\section{Method}

We propose \textbf{EC-RAG}, a training-free retrieval-augmented framework for large video-language models that organizes video content as structured event chains. As illustrated in Figure~\ref{fig:framework}, our pipeline comprises four stages: (i) \textbf{Query Decoupling}: The user query is decomposed into modality-specific retrieval requests for ASR, OCR, and DET. (ii) \textbf{CLIP-Variance Guided Multimodal Extraction}: Informative keyframes are selected through joint CLIP similarity and variance weighting, followed by parallel extraction of speech, text, and object information. (iii) \textbf{Temporal Event Chain Construction}: The video is segmented into temporal units and organized as a chain of semantically coherent events with explicit inter-event relations. (iv) \textbf{Evidence-augmented Multimodal Reasoning}: Query-guided localization identifies relevant events, and the LVLM generates answers grounded in the event chain context and multimodal evidence.

\textbf{Problem Formulation. } Given a video $V$ and a user query $Q$, a frame sampler extracts $N$ frames $\mathbf{F} = \{F_1, F_2, \ldots, F_N\}$. These frames are encoded by a visual encoder (e.g., \textbf{CLIP-L}~\cite{radford2021learning}) to obtain visual features $\mathbf{F}_v$. The standard LVLM inference can be formulated as:
\begin{equation}
O = \text{LVLM}(\mathbf{F}_v, Q)
\end{equation}
However, this direct approach struggles with long videos where critical information may be sparse across extended durations. \textbf{EC-RAG} addresses this limitation by constructing an \textbf{event chain (EC)} $\mathcal{E}$ that captures the temporal structure of video content, and performing \textbf{query-guided event localization} to retrieve relevant context before final reasoning:
\begin{equation}
O = \text{LVLM}(\mathbf{F}_v, Q, \mathcal{E}, \mathbf{A})
\end{equation}
where $\mathcal{E}$ denotes the temporal event chain and $\mathbf{A}$ represents the retrieved multimodal evidence. This formulation enables the model to reason over both visual frames and structured textual context derived from the video.

\subsection{Query Decoupling}

Before processing the video content, we first analyze the user query to understand what types of information are needed to answer it. Motivated by recent methods~\cite{lin2024video, luo2024video}, this query decoupling step separates the retrieval planning from the actual video processing, allowing the system to focus computational resources on relevant modalities. Upon receiving a user query, \textbf{EC-RAG} decouples it into structured retrieval requests without accessing video frames. The LVLM analyzes the query to determine what information sources are needed:
\begin{equation}
\mathbf{R} = \text{LVLM}(P_d, Q) = \{R_{\text{asr}}, R_{\text{ocr}}, R_{\text{det}}\}
\end{equation}
where $P_d$ is the decoupling prompt enhanced with few-shot examples that demonstrate how to extract retrieval targets from various question types. Each request may be \texttt{NULL} if the corresponding modality is unnecessary for the given query. Specifically, $R_{\text{asr}}$ specifies key phrases to retrieve from speech transcripts, which is particularly useful for questions about dialogue content or narration. $R_{\text{ocr}}$ defines keywords to search in on-screen text, targeting questions about titles, labels, or displayed information. $R_{\text{det}}$ indicates physical objects to detect in video frames, essential for visual grounding questions.

The decoupling prompt guides the LVLM to identify concrete retrieval targets through structured examples. For detection requests, we enforce that $R_{\text{det}}$ contains only physical entities rather than abstract concepts, ensuring compatibility with vision-based detection models. This constraint prevents the system from attempting to detect non-visual concepts like emotions or intentions.

\subsection{CLIP-Variance Guided Multimodal Extraction}

This stage selects informative keyframes and extracts multimodal information from the video in parallel.

\textbf{Keyframe Selection.} A critical challenge in video understanding is selecting which frames to process for detailed analysis. Uniform sampling may miss important moments, while processing all frames is computationally prohibitive. Rather than uniformly sampling frames or relying solely on semantic similarity, inspired by query-guided frame selection in prior work~\cite{cao2025tv, luo2024video}, we propose a \textbf{CLIP-variance (CV) guided selection} mechanism that jointly considers query relevance and visual informativeness. The intuition is that informative frames should both relate to the query content and contain rich visual details rather than static or redundant imagery. For each frame $F_t$, we compute a weighted score:
\begin{equation}
s_t = \text{sim}_{\text{CLIP}}(R_{\text{det}}, F_t) \cdot \sigma_t
\end{equation}
where $\text{sim}_{\text{CLIP}}(\cdot)$ denotes the CLIP similarity between the detection query and frame content. This term ensures that selected frames are semantically relevant to the objects or entities mentioned in the query. The second factor $\sigma_t$ is the \textbf{normalized visual variance} of frame $F_t$:
\begin{equation}
\sigma_t = \frac{\text{Var}(F_t)}{\max_j \text{Var}(F_j)}
\end{equation}

The variance term captures the visual complexity of each frame, computed from the pixel intensity distribution. Frames with higher variance typically contain more visual structure and details, while low-variance frames often correspond to static backgrounds, black screens, or uniform scenes that provide limited information. By multiplying these two factors, we favor frames that are both query-relevant and visually informative. Keyframes are selected by thresholding on the weighted scores:
\begin{equation}
\mathbf{F}_{\text{key}} = \{F_t \mid s_t \geq \tau_s\}
\end{equation}
where $\tau_s$ is a score threshold, with constraints on minimum and maximum frame counts to ensure adequate coverage across the video duration.

\textbf{ASR Extraction.} Audio information provides crucial context that is often unavailable through visual analysis alone, including dialogue, narration, and background audio cues.  Following recent retrieval-augmented video systems~\cite{cao2025tv, luo2024video}, we extract the audio track from the video and transcribe it using Whisper~\cite{radford2023robust}, a robust speech recognition model capable of handling diverse audio conditions. The resulting speech transcripts $T_{\text{asr}}$ are segmented by timestamps to preserve temporal alignment with video frames. These segments are then encoded using Contriever~\cite{izacard2021unsupervised}, a dense retrieval encoder trained for semantic similarity matching, producing embeddings that capture the semantic content of each transcript segment. The embeddings are indexed in a FAISS~\cite{johnson2019billion} database for efficient similarity search:
\begin{equation}
\text{DB}_{\text{asr}} \xleftarrow{\text{FAISS}} \mathbf{E}_{\text{asr}} = \text{Contriever}(T_{\text{asr}})
\end{equation}

\textbf{OCR Extraction.} On-screen text frequently contains valuable information such as titles, subtitles, labels, and other textual elements that complement visual and audio content. Current LVLMs often struggle to accurately recognize text within video frames, particularly for small or stylized fonts. To address this limitation, we employ EasyOCR~\cite{easyocr2023}, a dedicated text recognition model, to extract on-screen text from each sampled frame. Since OCR models may produce noisy outputs, we apply a confidence threshold to filter low-quality detections:
\begin{equation}
T_{\text{ocr}} = \{t \mid t = \text{EasyOCR}(F), \text{conf}(t) > \tau_{\text{ocr}}\}
\end{equation}
This filtering step removes unreliable text detections that could introduce noise into the retrieval process. The filtered texts are encoded and stored in a separate FAISS index following the same procedure as ASR:
\begin{equation}
\text{DB}_{\text{ocr}} \xleftarrow{\text{FAISS}} \mathbf{E}_{\text{ocr}} = \text{Contriever}(T_{\text{ocr}})
\end{equation}

\textbf{Object Detection with Scene Graph.} While LVLMs demonstrate strong capabilities in object recognition, they continue to face challenges in precise object counting, localization, and understanding spatial relationships between objects. These limitations often lead to hallucinations when answering questions that require fine-grained visual understanding. To provide more accurate object-level information, we employ APE~\cite{shen2024aligning}, an open-vocabulary object detection model that accepts textual prompts specifying which objects to detect. For the selected keyframes $\mathbf{F}_{\text{key}}$, APE localizes relevant objects based on the detection request:
\begin{equation}
D_t = \text{APE}(F_t, R_{\text{det}}), \quad F_t \in \mathbf{F}_{\text{key}}
\end{equation}

The raw detection outputs consisting of object categories and bounding box coordinates are not directly interpretable by language models. To bridge this gap, inspired by scene graph representations in visual reasoning~\cite{cao2025tv, luo2024video}, we transform the detection results into semantic \textbf{scene graph} descriptions expressed in natural language. For each keyframe, we generate three complementary types of information: (1) \textbf{Object Location} $A_{\text{loc}}$ converts bounding box coordinates into natural language spatial descriptions (e.g., ``left side of frame'', ``upper right corner''), making positional information accessible to the language model; (2) \textbf{Object Count} $A_{\text{cnt}}$ aggregates detection results to provide accurate counts for each object category, addressing the counting limitations of LVLMs; (3) \textbf{Object Relation} $A_{\text{rel}}$ computes and describes relative spatial relationships between detected objects (e.g., ``person A is to the left of person B''). The complete scene graph representation combines these components:

\begin{equation}
T_{\text{det}} = \text{SceneGraph}(A_{\text{loc}}, A_{\text{cnt}}, A_{\text{rel}})
\end{equation}

\subsection{Temporal Event Chain Construction}

Unlike frame-level retrieval methods that treat video as a collection of independent images, EC-RAG explicitly models the temporal structure of video content by constructing an event chain. This representation captures how events unfold over time and enables reasoning about temporal relationships.

\textbf{Video Segmentation.} Long videos contain extended sequences of content that cannot be processed as a single unit. We partition the video into non-overlapping temporal segments of fixed duration $\Delta$:
\begin{equation}
\mathcal{S} = \{S_i = [t_i, t_i + \Delta) \mid i = 0, 1, \ldots, \lfloor L/\Delta \rfloor\}
\end{equation}
where $L$ is the total video duration. Each segment $S_i$ serves as the basic unit for event extraction and is associated with the multimodal information (ASR, OCR, DET) extracted from frames within its time range. This segmentation balances granularity and efficiency: segments should be short enough to capture coherent events while being long enough to contain meaningful content.

\textbf{Event Extraction.} Raw multimodal data from each segment needs to be synthesized into a coherent semantic representation. For each segment $S_i$, we aggregate the corresponding ASR, OCR, and DET information and prompt the LVLM to extract a semantic event description that summarizes what happens in that temporal window:
\begin{equation}
e_i = \text{LVLM}(P_e, T_{\text{asr}}^{(i)}, T_{\text{ocr}}^{(i)}, T_{\text{det}}^{(i)})
\end{equation}
where $P_e$ is the event extraction prompt that instructs the model to integrate information across modalities. The prompt guides the LVLM to focus on the main action or occurrence while considering all available evidence. Each extracted event $e_i$ comprises three components: a description that concisely summarizes what happens in this segment by integrating speech, text, and visual information; a type that classifies the event as action, dialogue, scene transition, or informational content; and a list of participants identifying key entities (people, objects) involved in the event.

\textbf{Event Chain Assembly.} Individual events are connected to form a coherent narrative structure. The extracted events are organized into a \textbf{temporal event chain} $\mathcal{E} = \{e_0, e_1, \ldots, e_n\}$ with explicit temporal ordering that reflects the video's chronological structure. We establish predecessor-successor relations between adjacent events:
\begin{equation}
e_i \xrightarrow{\text{temporal}} e_{i+1}, \quad \forall i \in [0, n-1]
\end{equation}

The complete event chain provides a structured overview of the video's narrative progression, capturing how events unfold and connect across time. This representation enables the model to understand not just individual moments but also the broader context and temporal dynamics of the video content.

\subsection{Evidence-augmented Multimodal Reasoning}

With the event chain constructed, EC-RAG performs query-guided retrieval to identify relevant events and gather supporting evidence for answer generation.

\textbf{Query-guided Event Localization.} Not all events in a video are relevant to a given query. We identify the most relevant events through semantic similarity matching between the query and event descriptions. Each event description is encoded using \textbf{Contriever} and compared against the query embedding using cosine similarity:
\begin{equation}
\text{sim}_i = \cos\bigl(\text{Contriever}(Q),\; \text{Contriever}(e_i.\text{desc})\bigr)
\end{equation}

This text-to-text matching is more appropriate than image-text matching (e.g., CLIP) since both the query and event descriptions are textual. Events exceeding a relevance threshold $\tau_{\text{loc}}$ are selected as located events:
\begin{equation}
\mathcal{E}_{\text{loc}} = \{e_i \mid \text{sim}_i > \tau_{\text{loc}}\}
\end{equation}

The located events represent the temporal regions most likely to contain information needed to answer the query.

\textbf{Evidence Retrieval.} While event descriptions provide high-level summaries, detailed evidence from the original multimodal sources may be necessary for accurate reasoning. For each located event $e_i \in \mathcal{E}_{\text{loc}}$, we retrieve the associated multimodal evidence from the constructed databases. Using the retrieval requests $\mathbf{R}$ generated during query decoupling, we query the ASR and OCR databases to obtain relevant text segments:
\begin{equation}
A_{\text{asr}} = \text{Retrieve}(\text{DB}_{\text{asr}}, R_{\text{asr}}, \tau), \quad A_{\text{ocr}} = \text{Retrieve}(\text{DB}_{\text{ocr}}, R_{\text{ocr}}, \tau)
\end{equation}
The DET evidence is collected from the scene graphs of keyframes falling within the located events' time ranges:
\begin{equation}
A_{\text{det}} = \{T_{\text{det}}^{(t)} \mid F_t \in \mathbf{F}_{\text{key}},\; t \in e_i.\text{span},\; e_i \in \mathcal{E}_{\text{loc}}\}
\end{equation}
where the retrieval function returns all database entries with similarity scores exceeding threshold $\tau$. The DET evidence $A_{\text{det}}$ is directly obtained from the scene graph descriptions of keyframes within the located events' time ranges, providing detailed object-level information for visual grounding.

\textbf{Context Construction and Answer Generation.} The final reasoning step integrates multiple levels of context to provide the LVLM with comprehensive information. The context combines three complementary components: (1) \textbf{Event Chain Summary} provides a global overview of all events in the video, giving the model awareness of the full narrative arc even for content not directly relevant to the query; (2) \textbf{Located Events} contain detailed descriptions of query-relevant events along with their temporal predecessors, enabling understanding of causal relationships and temporal context; (3) \textbf{Multimodal Evidence} includes the raw ASR, OCR, and DET information retrieved for the located events, providing specific details that support accurate reasoning. The complete context is assembled as:
\begin{equation}
C = \text{Concat}(\text{Summary}(\mathcal{E}), \text{Detail}(\mathcal{E}_{\text{loc}}), A_{\text{asr}}, A_{\text{ocr}}, A_{\text{det}})
\end{equation}

Finally, the LVLM generates the answer by jointly reasoning over the video frames, user query, and constructed context:
\begin{equation}
O = \text{LVLM}(\mathbf{F}_v, Q, C)
\end{equation}

\begin{table*}[!ht]
\centering
\caption{Performance on the Video-MME~\cite{fu2025video} benchmark. \textbf{\#Text} indicates the volume of retrieved textual context. EC-RAG is integrated into four open-source LVLMs spanning 8--32 input frames.}
\label{tab:videomme}
\small
\begin{tabular}{l|ccc|cccc|c}
\toprule
\textbf{Model} & \textbf{\#Text} & \textbf{LLM Params} & \textbf{Frames} & \textbf{Short} & \textbf{Medium} & \textbf{Long} & \textbf{Overall} & \textbf{Gain} \\
\midrule
\multicolumn{9}{c}{\textit{Proprietary LVLMs}} \\
\midrule
GPT-4o~\cite{openai2024gpt4o}          & -- & -- & 384      & 80.0 & 70.3 & 65.3 & 71.9 & -- \\
Gemini-1.5-Pro~\cite{team2024gemini}    & -- & -- & 0.5\,fps & 81.7 & 74.3 & 67.4 & 75.0 & -- \\
\midrule
\multicolumn{9}{c}{\textit{Open-Source LVLMs}} \\
\midrule
Video-LLaVA~\cite{lin2024video}                 & --   & 7B  & 8  & 44.6 & 38.3 & 35.8 & 39.6 & -- \\
\rowcolor{blue!10}
Video-LLaVA + EC-RAG                             & 2.0K & 7B  & 8  & 50.1 & 44.6 & 43.4 & 46.0 & \textcolor{red}{+6.4} \\
LLaVA-NeXT-Video~\cite{zhang2024llavanextvideo}  & --   & 7B  & 16 & 49.4 & 43.0 & 36.7 & 43.0 & -- \\
\rowcolor{blue!10}
LLaVA-NeXT-Video + EC-RAG                        & 2.0K & 7B  & 16 & 55.8 & 53.2 & 52.5 & 53.8 & \textcolor{red}{+10.8} \\
Long-LLaVA~\cite{long-llava-qwen2-7b}           & --   & 7B  & 32 & 60.3 & 51.4 & 44.1 & 52.0 & -- \\
\rowcolor{blue!10}
Long-LLaVA + EC-RAG                              & 1.9K & 7B  & 32 & 67.6 & 60.3 & 60.0 & 62.6 & \textcolor{red}{+10.6} \\
LLaVA-Video~\cite{zhang2024llava}               & --   & 7B  & 32 & 75.7 & 59.6 & 52.1 & 62.4 & -- \\
LLaVA-Video + Video-RAG~\cite{luo2024video}      & 2.0K & 7B  & 32 & 70.9 & 63.2 & 59.0 & 64.3 & \textcolor{red}{+1.9} \\
LLaVA-Video + TV-RAG~\cite{cao2025tv}            & 2.0K   & 7B  & 32 & 72.4 & 64.1 & 59.8 & 65.4 & \textcolor{red}{+3.0} \\
\rowcolor{blue!10}
LLaVA-Video + EC-RAG                             & 2.0K & 7B  & 32 & 77.8 & 64.3 & 60.4 & 67.5 & \textcolor{red}{+5.1} \\
LLaVA-Video~\cite{zhang2024llava}               & --   & 72B & 32 & 78.0 & 63.7 & 59.6 & 67.1 & -- \\
\rowcolor{blue!10}
LLaVA-Video + EC-RAG                             & 2.1K & 72B & 32 & \textbf{81.9} & \textbf{72.4} & \textbf{72.7} & \textbf{75.7} & \textcolor{red}{+8.6} \\
\bottomrule
\end{tabular}
\end{table*}

\section{Experiments}

\subsection{Datasets}

We assess EC-RAG on three widely adopted long-video benchmarks that span diverse durations, domains, and reasoning demands.

\textbf{Video-MME}~\cite{fu2025video} offers a systematic testbed for evaluating multimodal comprehension over real-world video content.
Its 900 videos are grouped by length into \textit{short} ($<$\,2\,min), \textit{medium} (4--15\,min), and \textit{long} (30--60\,min) subsets, yielding 2{,}700 multiple-choice questions across 12 task types---ranging from object recognition and counting to temporal reasoning and OCR---and 30 sub-categories drawn from six broad domains including knowledge, sports, and film.
This breadth makes Video-MME particularly suitable for stress-testing whether a retrieval-augmented pipeline can supply the right modality of evidence for each question type.

\textbf{MLVU}~\cite{zhou2024mlvu} assembles nine evaluation tasks from videos of 3\,min to 2\,h (mean $\approx$\,12\,min), requiring models to sustain reasoning over extended temporal ranges

\textbf{LongVideoBench}~\cite{wu2024longvideobench} focuses on multimodal retrieval and compositional reasoning within lengthy footage.
It provides 6{,}678 human-written multiple-choice questions spanning 17 thematic categories, testing a model's ability to pinpoint and integrate evidence scattered across long video timelines.

\subsection{Implementation Details}

All experiments were conducted on NVIDIA RTX 4090 and A800 80\,GB GPUs.
We adopt \textbf{LLaVA-Video-7B}~\cite{zhang2024llava} as the primary LVLM backbone, with three additional 7B architectures and the 72B variant evaluated in Table~\ref{tab:videomme} to verify cross-architecture and cross-scale generalizability.
Detection requests produced during query decoupling are post-filtered with spaCy to retain only concrete, CLIP-responsive entities.
Both the CLIP similarity threshold and the FAISS~\cite{johnson2019billion} retrieval threshold are fixed at $\tau = 0.3$, with \texttt{IndexFlatIP} as the similarity backend.
Videos are segmented into 30-second windows for event chain construction.
Speech is transcribed by Whisper-Large~\cite{radford2023robust}, on-screen text is recognized by EasyOCR~\cite{easyocr2023}, open-vocabulary detection is handled by APE~\cite{shen2024aligning}, and dense text retrieval is performed by Contriever~\cite{izacard2021unsupervised}.
Our ablation suite is built around LLaVA-Video-7B at 32\,frames, whose moderate context window makes it a practical testbed for probing the effect of individual components and hyper-parameters.


\textbf{Video-MME.}
To verify that EC-RAG generalizes across architectures, model scales, and frame budgets, we integrate it into four open-source LVLMs spanning 8--32 input frames, as reported in Table~\ref{tab:videomme}. EC-RAG yields consistent improvements for every backbone, with gains most pronounced on \textit{long} videos where uniform sampling alone cannot cover the query-relevant content. The benefit is especially evident on frame-constrained backbones, confirming that event-chain retrieval compensates effectively for sparse visual input. The pipeline injects approximately 2\,K tokens of supplementary textual context (the \textbf{\#Text} column), comparable to roughly 14 extra keyframes in payload, which complement the visual input and help anchor temporal reasoning.

Scaling to the 72B LLaVA-Video, EC-RAG achieves 75.7\% overall, surpassing both GPT-4o~\cite{openai2024gpt4o} and Gemini-1.5-Pro~\cite{team2024gemini}, showing that structured event-chain retrieval enables open-source models to outperform proprietary systems without additional pretraining. EC-RAG also consistently outperforms Video-RAG~\cite{luo2024video} and TV-RAG~\cite{cao2025tv} under the same backbone setting, as event-level temporal localization provides more precise context than frame-level retrieval.

\begin{table}[t]
\centering
\caption{Overall accuracy on the multiple-choice split of the MLVU~\cite{zhou2024mlvu} benchmark. EC-RAG achieves the best result among all 7B-scale models.}
\label{tab:mlvu}
\begin{tabular}{l|cc|c}
\toprule
\textbf{Model} & \textbf{\#Params} & \textbf{Frames} & \textbf{Overall} \\
\midrule
\multicolumn{4}{c}{\textit{Proprietary LVLMs}} \\
\midrule
GPT-4o~\cite{openai2024gpt4o} & -- & 0.5\,fps & 64.6 \\
\midrule
\multicolumn{4}{c}{\textit{Open-Source LVLMs}} \\
\midrule
Video-CCAM~\cite{fei2024video}   & 14B   & 96  & 63.1 \\
Video-XL~\cite{shu2025video}       & 7B    & 256 & 64.9 \\
Aria~\cite{li2024aria}               & 25.3B & 256 & 70.6 \\
LLaVA-Video~\cite{zhang2024llava}    & 7B    & 64  & 70.8 \\
Oryx-1.5~\cite{liu2024oryx}          & 32B   & 128 & 72.3 \\
LLaVA-Video + Video-RAG~\cite{luo2024video} & 7B & 64 & 72.4 \\
LLaVA-Video + TV-RAG~\cite{cao2025tv}      & 7B & 64 & 72.6 \\
\rowcolor{blue!10}
LLaVA-Video + EC-RAG                  & 7B & 64 & \textbf{72.9} \\
\bottomrule
\end{tabular}
\end{table}

\begin{figure*}[h]
  \centering
  \includegraphics[width=\linewidth]{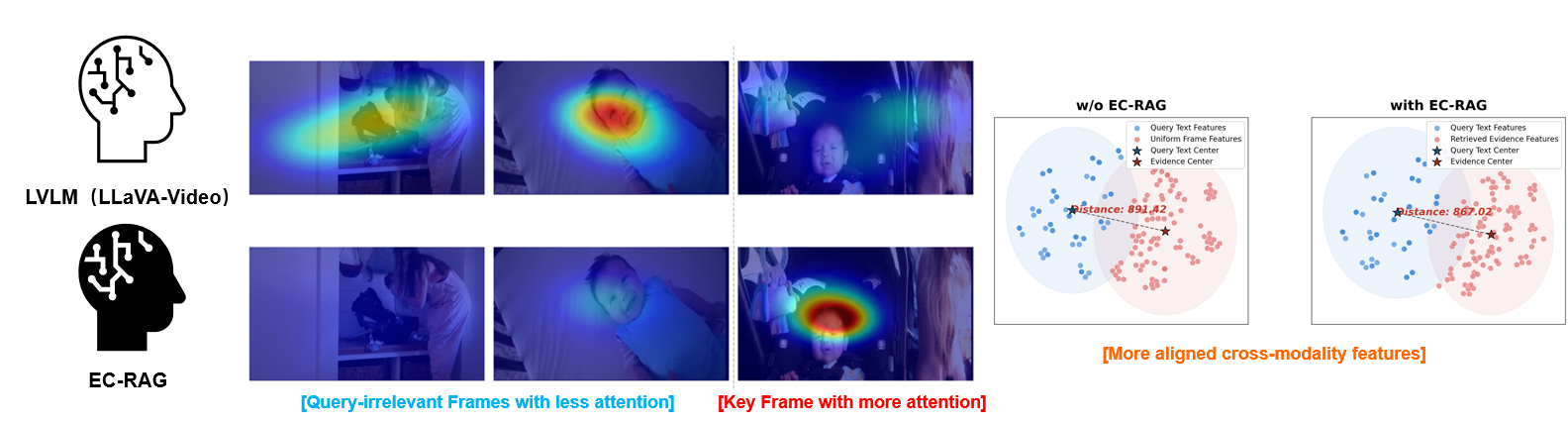}
  \caption{Left: Grad-CAM heatmaps on query-relevant and query-irrelevant frames for the baseline and EC-RAG. Right: t-SNE projection of query, vision, and text features, showing tighter cross-modal alignment when EC-RAG is applied.}
  \Description{Grad-CAM attention heatmaps and t-SNE feature projections comparing baseline and EC-RAG cross-modal alignment.}
  \label{fig:gradcam_tsne}
\end{figure*}


\textbf{MLVU.}
Table~\ref{tab:mlvu} reports performance on the MLVU multiple-choice task.
EC-RAG lifts LLaVA-Video-7B from 70.8 to \textbf{72.9}, outperforming both Video-RAG and TV-RAG on the same backbone.
Notably, this 7B model with only 64 frames also surpasses Oryx-1.5~\cite{liu2024oryx}, a 32B model operating on 128 frames,  highlighting the efficiency of event-chain retrieval over brute-force scaling.

\textbf{LongVideoBench.}
As shown in Table~\ref{tab:lvb}, EC-RAG also achieves the best result on LongVideoBench, reaching \textbf{59.5} with LLaVA-Video-7B at 64 frames.
This benchmark requires retrieving and composing evidence distributed across long timelines, a scenario that directly benefits from EC-RAG's temporal event chain design.
Compared with Video-RAG and TV-RAG under the same setting, EC-RAG provides a more substantial boost over the unaugmented baseline, indicating that event-level localization is more effective than flat retrieval when evidence is temporally scattered.
\begin{table}[t]
\centering
\caption{Performance on the Long\-Video\-Bench~\cite{wu2024longvideobench} validation set.}
\label{tab:lvb}
\begin{tabular}{l|cc|c}
\toprule
\textbf{Model} & \textbf{\#Params} & \textbf{Frames} & \textbf{Overall} \\
\midrule
VideoChat2-Mistral~\cite{li2025videochat}       & 7B  & 8  & 39.3 \\
ShareGPT4Video~\cite{chen2024sharegpt4video}   & 7B  & 8  & 39.7 \\
LLaVA-Next-Mistral~\cite{zhang2024llavanextvideo} & 7B & 8 & 49.1 \\
PLLaVA~\cite{xu2024pllava}                     & 34B & 16 & 53.2 \\
LLaVA-Video~\cite{zhang2024llava}              & 7B  & 64 & 56.6 \\
LLaVA-Video + Video-RAG~\cite{luo2024video}    & 7B  & 64 & 58.7 \\
LLaVA-Video + TV-RAG~\cite{cao2025tv}          & 7B  & 64 & 58.8 \\
\rowcolor{blue!10}
LLaVA-Video + EC-RAG                    & 7B  & 64 & \textbf{59.5} \\
\bottomrule
\end{tabular}
\end{table}

\subsection{Ablation Studies}
\label{sec:ablation}
\begin{figure*}[h]
  \centering
  \includegraphics[width=\linewidth]{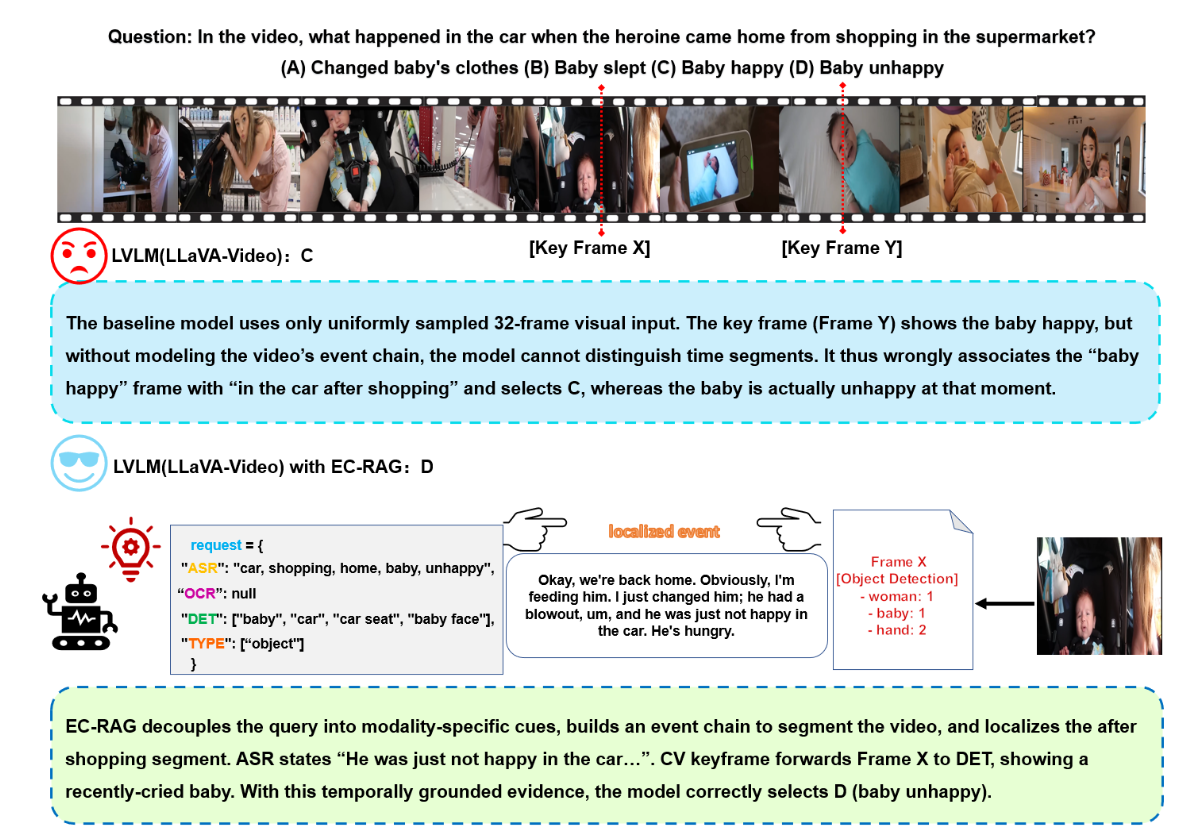}
  \caption{Qualitative comparison between the baseline LLaVA-Video and EC-RAG on a Video-MME example. The baseline confuses temporally distinct segments, whereas EC-RAG localizes the query-relevant event through its event chain and aggregates ASR and DET evidence to arrive at the correct answer.}
  \Description{A qualitative case study showing how EC-RAG correctly answers a temporal reasoning question by localizing the relevant event and aggregating multimodal evidence.}
\label{fig:qualitative}
\end{figure*}
To quantify each component's contribution, we ablate EC-RAG on Video-MME with LLaVA-Video-7B at 32\,frames (Table~\ref{tab:ablation}).
Among all modules, ASR has the greatest impact: removing it causes the most significant performance degradation, particularly on medium and long videos, confirming that speech transcription is the most informative auxiliary modality for video QA.
The event chain module ranks second in importance, as disabling temporal localization leads to less targeted evidence retrieval.
CLIP-variance guided keyframe selection, OCR, and DET each provide meaningful but comparatively smaller gains, reflecting their complementary roles in selecting informative frames, capturing on-screen text, and grounding physical entities respectively.

\begin{table}[t]
\centering
\caption{Component ablation of EC-RAG on Video-MME with LLaVA-Video-7B.}
\label{tab:ablation}
\resizebox{\linewidth}{!}{
\begin{tabular}{ccccc|cccc}
\toprule
\textbf{ASR} & \textbf{OCR} & \textbf{DET} & \textbf{EC} & \textbf{CV} & \textbf{Short} & \textbf{Medium} & \textbf{Long} & \textbf{Overall} \\
\midrule
$\times$   & \checkmark & \checkmark & \checkmark & \checkmark & 74.8 & 60.3 & 54.7 & 63.3 \\
\checkmark & $\times$   & \checkmark & \checkmark & \checkmark & 76.5 & 62.8 & 59.9 & 66.4 \\
\checkmark & \checkmark & $\times$   & \checkmark & \checkmark & 77.1 & 62.9 & 59.2 & 66.4 \\
\checkmark & \checkmark & \checkmark & $\times$   & \checkmark & 77.0 & 61.6 & 58.4 & 65.7 \\
\checkmark & \checkmark & \checkmark & \checkmark & $\times$   & 76.4 & 63.5 & 58.9 & 66.3 \\
\rowcolor{blue!10}
\checkmark & \checkmark & \checkmark & \checkmark & \checkmark & \textbf{77.8} & \textbf{64.3} & \textbf{60.4} & \textbf{67.5} \\
\bottomrule
\end{tabular}
}
\end{table}
\subsection{Qualitative Analysis}

Figure~\ref{fig:qualitative} presents a representative case from Video-MME in which the baseline LLaVA-Video fails but EC-RAG succeeds.
The question asks what happened in the car after the heroine returned from shopping.
With 32 uniformly sampled frames, the baseline captures a frame showing the baby in a cheerful state and, lacking any temporal context to disambiguate it from other segments, incorrectly selects option~C (``Baby happy'').

EC-RAG addresses this through its pipeline: the query decoupling stage extracts modality-specific retrieval cues (e.g., ASR keywords ``car, shopping, unhappy'' and DET targets ``baby, car seat'').
The event chain then localizes the relevant temporal segment, whose ASR transcript explicitly states ``He was just not happy in the car.''
Simultaneously, CLIP-variance guided keyframe selection identifies a visually informative frame that is forwarded to the object detector, revealing a baby being fed with a recently-cried expression.
Equipped with this temporally grounded multimodal evidence, the model correctly selects option~D (``Baby unhappy'').

Figure~\ref{fig:gradcam_tsne} further illustrates the effect of EC-RAG through Grad-CAM heatmaps and t-SNE projections.
The Grad-CAM visualizations compare the attention distribution of the baseline and EC-RAG on both a query-relevant and a query-irrelevant frame.
With the supplementary textual evidence provided by EC-RAG, the model's attention concentrates more tightly on the semantically relevant regions (e.g., the baby and car seat), whereas the baseline distributes attention more diffusely.
The t-SNE plot projects the query, vision, and text features into a shared 2D space, showing that EC-RAG's auxiliary tokens pull the vision and query representations closer together, resulting in more aligned cross-modal features that facilitate accurate reasoning.
\section{Conclusion}

We presented EC-RAG, a training-free retrieval-augmented generation framework that structures long-video understanding around temporal event chains. By decomposing videos into semantically coherent events through the fusion of ASR, OCR, and open-vocabulary detection, and localizing query-relevant segments before injecting multimodal evidence into the LVLM prompt, EC-RAG consistently improves diverse backbones across three benchmarks---Video-MME, MLVU, and LongVideoBench---outperforming existing RAG-based methods under every tested setting. Notably, with the 72B LLaVA-Video backbone, EC-RAG surpasses both GPT-4o and Gemini-1.5-Pro on Video-MME using a fully open-source pipeline, demonstrating that structured event-level retrieval enables open-source models to rival and even surpass proprietary systems without additional training.

Beyond the benchmarks evaluated in this work, we believe the event-chain paradigm holds broader potential for other video-centric tasks such as video summarization, temporal grounding, and multi-turn video dialogue, where preserving temporal structure and aggregating cross-modal evidence are equally critical. The training-free nature of EC-RAG also makes it readily applicable to emerging LVLMs as they continue to evolve, offering a flexible and scalable augmentation strategy.

\bibliographystyle{ACM-Reference-Format}
\bibliography{references}

\newpage
\appendix

\section*{Supplementary Material}

\section{Additional Ablation Studies}
\label{sec:appendix_ablation}

\begin{figure}[t]
  \centering
  \includegraphics[width=\linewidth]{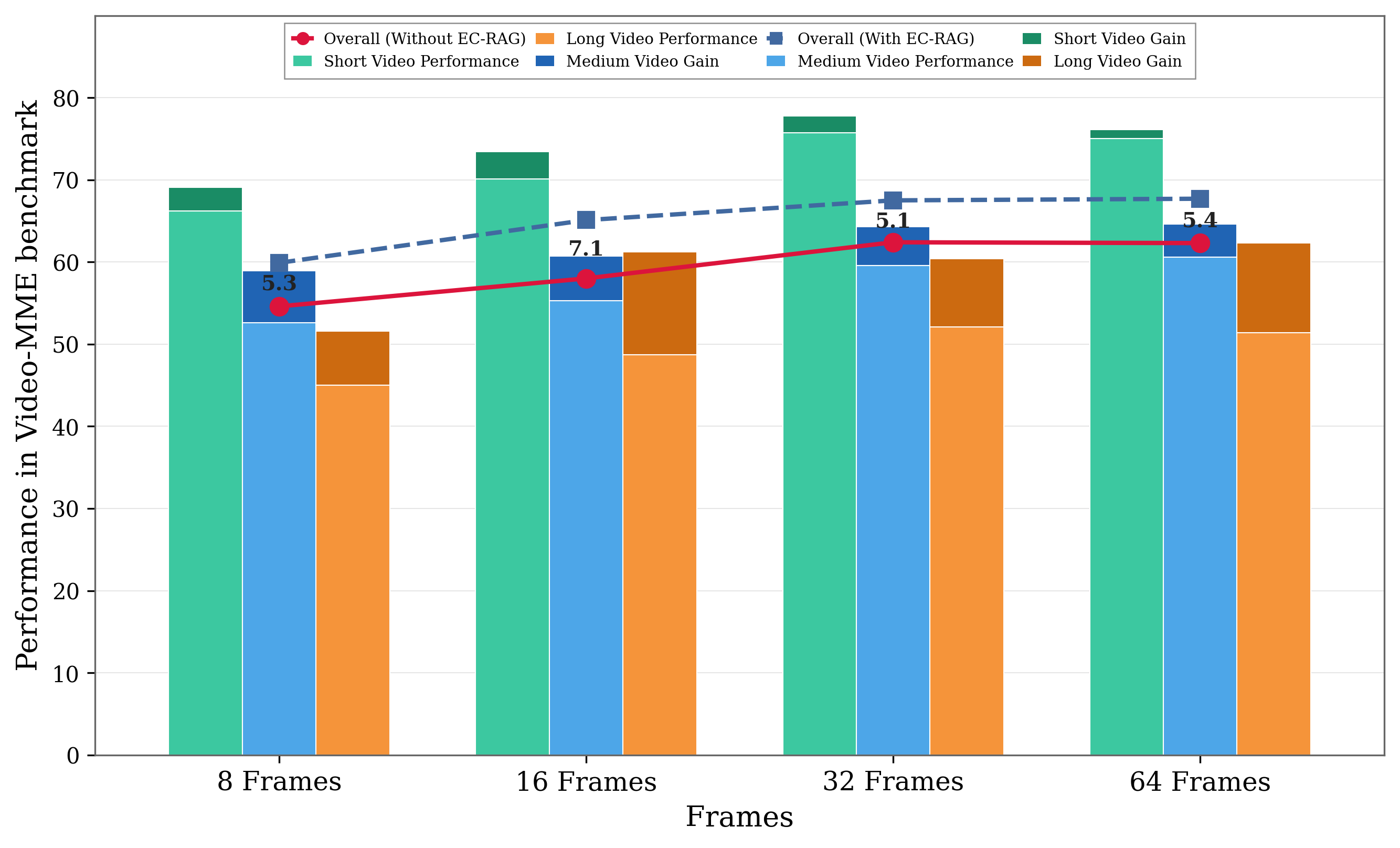}
  \caption{Accuracy of LLaVA-Video-7B~\cite{zhang2024llava} with and without EC-RAG under different frame budgets on Video-MME~\cite{fu2025video}. EC-RAG provides consistent gains at all budgets, with the largest improvement at 16 frames.}
  \Description{Bar chart comparing baseline and EC-RAG accuracy across 8, 16, and 32 frames on short, medium, and long videos.}
  \label{fig:frame_budget}
\end{figure}

\textbf{Frame Sampling Budget.}
We investigate the sensitivity of EC-RAG to the number of sampled frames by evaluating LLaVA-Video-7B~\cite{zhang2024llava} under three frame budgets: 8, 16, and 32 on Video-MME~\cite{fu2025video}.
As shown in Figure~\ref{fig:frame_budget}, EC-RAG consistently outperforms the baseline at every frame budget across all duration categories.
Notably, the baseline shows diminishing returns as the frame count increases—particularly on long videos, where doubling from 16 to 32 frames yields only marginal improvement, suggesting that simply adding more visual input cannot compensate for the lack of temporal and textual context.
In contrast, EC-RAG maintains substantial gains at all budgets, with the largest absolute improvement observed at 16 frames, indicating that event-chain retrieval is especially effective when the visual input is moderately sparse and the model most needs supplementary evidence to bridge temporal gaps.

\textbf{Retrieval Threshold.}
Table~\ref{tab:threshold} investigates the effect of the retrieval similarity threshold $\tau$ on both efficiency and accuracy. Lower thresholds ($\tau{=}0.0$--$0.1$) retain nearly all retrieved text, injecting up to 4.4\,K supplementary tokens and pushing inference time to over 46\,s per question, yet only achieve 64.1--64.7\% overall. As $\tau$ increases, the token budget and latency decrease steadily; at $\tau{=}0.3$, EC-RAG processes 2.4\,K tokens in 18\,s while reaching the peak overall accuracy of 65.1\%, with the best short-video (73.4\%) and long-video (61.2\%) scores. This non-monotonic pattern indicates that unfiltered retrieval introduces noisy evidence that can mislead the model. Beyond $\tau{=}0.3$, accuracy degrades as useful context is progressively pruned: $\tau{=}0.5$ drops to 60.8\% and $\tau{=}1.0$---which effectively discards all retrieved text (0.0\,K tokens)---falls to 58.9\%. We therefore adopt $\tau{=}0.3$ as the default, which reduces latency by roughly 65\% compared to unfiltered retrieval while maximizing accuracy.

\textbf{Cross-Backbone RAG Comparison.}
Table~\ref{tab:compare_rag} compares EC-RAG with Video-RAG~\cite{luo2024video} and TV-RAG~\cite{cao2025tv} under two additional backbone settings. On Video-LLaVA-7B~\cite{lin2024video} with 8 frames, all three RAG methods substantially improve over the baseline (39.6\%), with EC-RAG achieving the highest overall accuracy of 46.0\%, outperforming TV-RAG by +0.5 and Video-RAG by +1.0. The advantage is most pronounced on long videos (+0.8 over TV-RAG), where event-chain retrieval provides more temporally precise context than frame-level or segment-level approaches. On LongVA-7B~\cite{zhang2024long} with 32 frames, the same trend holds: EC-RAG reaches 62.8\% overall, surpassing TV-RAG (61.7\%) and Video-RAG (60.1\%). EC-RAG consistently leads across all duration categories under both backbones, confirming that structured event-level retrieval generalizes across different LVLMs and frame budgets.
\begin{table}[t]
\centering
\caption{Performance with different retrieval thresholds on Video-MME~\cite{fu2025video} with LLaVA-Video-7B~\cite{zhang2024llava}.}
\label{tab:threshold}
\begin{tabular}{c|cc|cccc}
\toprule
$\tau$ & \textbf{\#Token} & \textbf{Time} & \textbf{Short} & \textbf{Medium} & \textbf{Long} & \textbf{Overall} \\
\midrule
0.0 & 4.4K & 52s & 72.4 & 61.6 & 60.0 & 64.7 \\
0.1 & 4.0K & 46s & 71.8 & 60.5 & 59.9 & 64.1 \\
0.2 & 3.5K & 23s & 70.3 & 60.4 & 57.8 & 62.8 \\
\rowcolor{blue!10}
0.3 & 2.4K & 18s & \textbf{73.4} & \textbf{60.7} & \textbf{61.2} & \textbf{65.1} \\
0.4 & 1.6K & 13s & 70.1 & 60.6 & 57.5 & 62.7 \\
0.5 & 1.2K & 12s & 68.9 & 58.2 & 55.3 & 60.8 \\
1.0 & 0.0K & 10s & 66.5 & 56.4 & 53.7 & 58.9 \\
\bottomrule
\end{tabular}
\end{table}

\begin{table}[t]
\centering
\caption{Comparison with other RAG methods across different backbones on Video-MME~\cite{fu2025video}.}
\label{tab:compare_rag}
\resizebox{\linewidth}{!}{
\begin{tabular}{l|cc|cccc}
\toprule
\textbf{Model} & \textbf{Params} & \textbf{Frames} & \textbf{Short} & \textbf{Medium} & \textbf{Long} & \textbf{Overall} \\
\midrule
X=Video-LLaVA~\cite{lin2024video}          & 7B & 8  & 44.6 & 38.3 & 35.8 & 39.6 \\
X+Video-RAG~\cite{luo2024video}            & 7B & 8  & 49.5 & 43.0 & 42.5 & 45.0 \\
X+TV-RAG~\cite{cao2025tv}                  & 7B & 8  & 49.7 & 44.3 & 42.6 & 45.5 \\
\rowcolor{blue!10}
X+EC-RAG                                   & 7B & 8  & \textbf{50.1} & \textbf{44.6} & \textbf{43.4} & \textbf{46.0} \\
\midrule
X=LongVA~\cite{zhang2024long}              & 7B & 32 & 60.9 & 49.3 & 44.0 & 51.4 \\
X+Video-RAG~\cite{luo2024video}            & 7B & 32 & 65.4 & 59.1 & 55.7 & 60.1 \\
X+TV-RAG~\cite{cao2025tv}                  & 7B & 32 & 66.2 & 62.1 & 58.7 & 61.7 \\
\rowcolor{blue!10}
X+EC-RAG                                   & 7B & 32 & \textbf{67.1} & \textbf{62.5} & \textbf{58.9} & \textbf{62.8} \\
\bottomrule
\end{tabular}
}
\end{table}

\section{Task-Type Breakdown}
\label{sec:appendix_task}
\begin{table*}[!ht]
\centering
\caption{Per-task-type accuracy (\%) on Video-MME~\cite{fu2025video} with LLaVA-Video-7B~\cite{zhang2024llava} as the backbone. \textbf{Gain} denotes the absolute improvement over the baseline. Best results per frame setting are in \textbf{bold}.}
\label{tab:task_type}
\resizebox{\textwidth}{!}{%
\begin{tabular}{l|c|cccccccccccc|c|c}
\toprule
\textbf{Method} & \textbf{Frames}
  & \textbf{TmpR} & \textbf{ActR} & \textbf{ActRec} & \textbf{AttrP}
  & \textbf{SpaP} & \textbf{SpaR} & \textbf{TmpP} & \textbf{InfoS}
  & \textbf{OCR} & \textbf{ObjR} & \textbf{ObjRec} & \textbf{Count}
  & \textbf{AVG} & \textbf{Gain} \\
\midrule
LLaVA-Video~\cite{zhang2024llava}          & 8  & 41.8 & 47.4 & 55.6 & 69.4 & 59.3 & 71.4 & 56.4 & 70.0 & 48.9 & 51.1 & 58.5 & 37.7 & 54.6 & -- \\
\rowcolor{blue!8}
LLaVA-Video + EC-RAG  & 8  & \textbf{46.9} & \textbf{49.8} & \textbf{55.9} & \textbf{73.9} & \textbf{59.6} & \textbf{80.4} & \textbf{60.0} & \textbf{79.6} & \textbf{56.8} & \textbf{60.6} & \textbf{63.6} & \textbf{40.3} & \textbf{60.2} & +5.6 \\
\midrule
LLaVA-Video~\cite{zhang2024llava}          & 16 & 45.2 & 51.2 & 57.5 & 68.0 & 57.4 & 73.2 & \textbf{69.1} & 74.0 & 50.4 & 55.4 & 64.1 & \textbf{41.8} & 58.0 & -- \\
\rowcolor{blue!8}
LLaVA-Video + EC-RAG  & 16 & \textbf{51.4} & \textbf{64.9} & \textbf{62.6} & \textbf{78.4} & \textbf{61.1} & \textbf{80.4} & \textbf{69.6} & \textbf{83.9} & \textbf{59.0} & \textbf{66.1} & \textbf{67.2} & \textbf{42.3} & \textbf{65.8} & +7.8 \\
\bottomrule
\end{tabular}%
}
\end{table*}

Table~\ref{tab:task_type} presents a fine-grained breakdown of accuracy across all 12 task types on Video-MME under 8-frame and 16-frame settings. With 8 frames, EC-RAG improves the average by +5.6 points (54.6\%$\to$60.2\%), with the largest gains on Information Synopsis (+9.6), Spatial Reasoning (+9.0), and OCR (+7.9)---tasks that heavily rely on textual or spatial cues where event-chain evidence provides direct support. At 16 frames, the average gain further widens to +7.8 points (58.0\%$\to$65.8\%), with Action Reasoning seeing the most dramatic improvement (+13.7) followed by Attribute Perception (+10.4) and Information Synopsis (+9.9). EC-RAG achieves the best score on 10 out of 12 task types at 8 frames and 12 out of 12 at 16 frames, demonstrating that the event-chain retrieval complements visual features comprehensively across diverse question types rather than benefiting only a narrow subset.
\begin{figure*}[!ht]
  \centering
  \includegraphics[width=\linewidth]{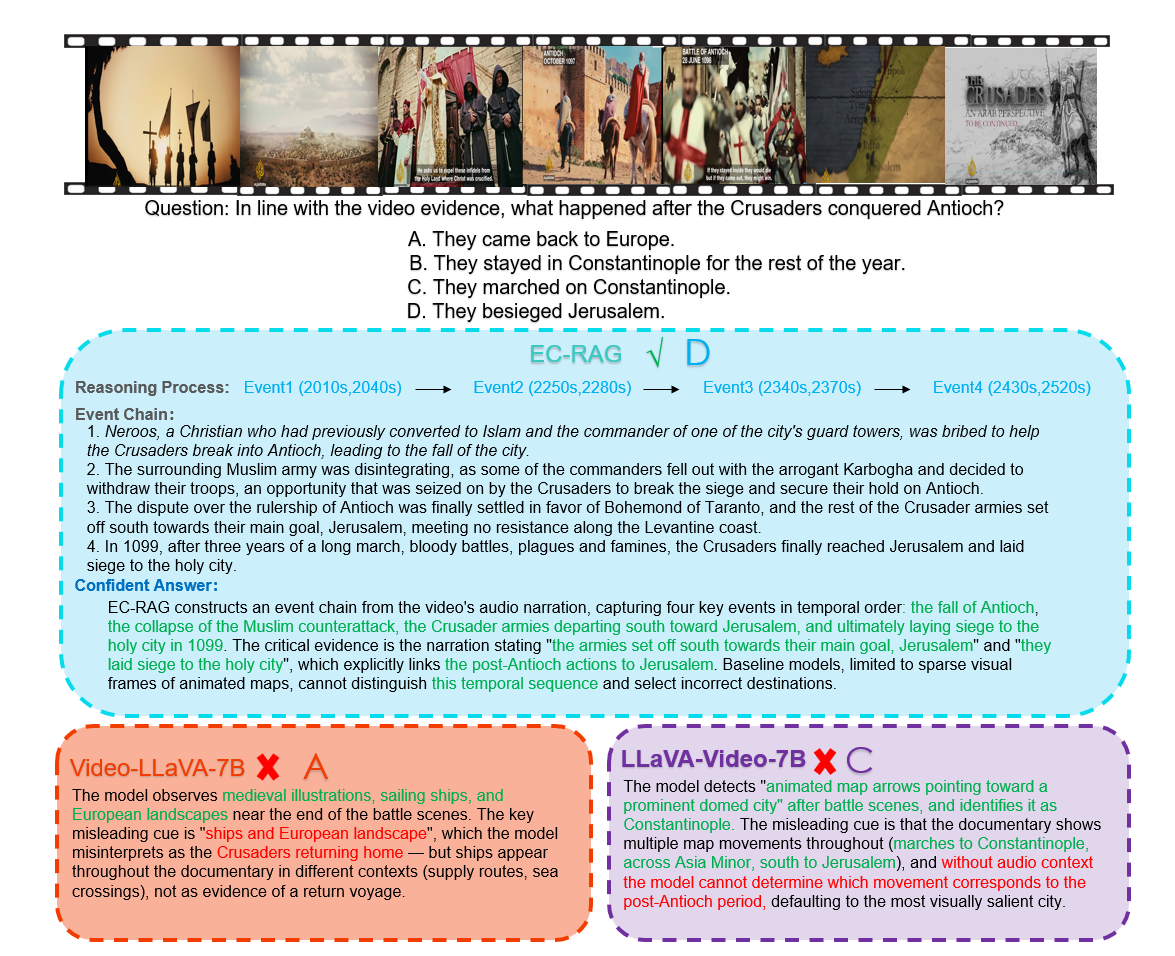}
  \caption{Qualitative example 1: a history documentary where the question requires reasoning about the temporal order of events after a key battle. The baseline confuses visually similar city depictions, while EC-RAG traces the event chain through ASR narration to identify the correct subsequent event.}
  \Description{Qualitative comparison between baseline and EC-RAG on a history documentary temporal reasoning question.}
  \label{fig:qualitative_case1}
\end{figure*}

\begin{figure*}[!ht]
  \centering
  \includegraphics[width=\linewidth]{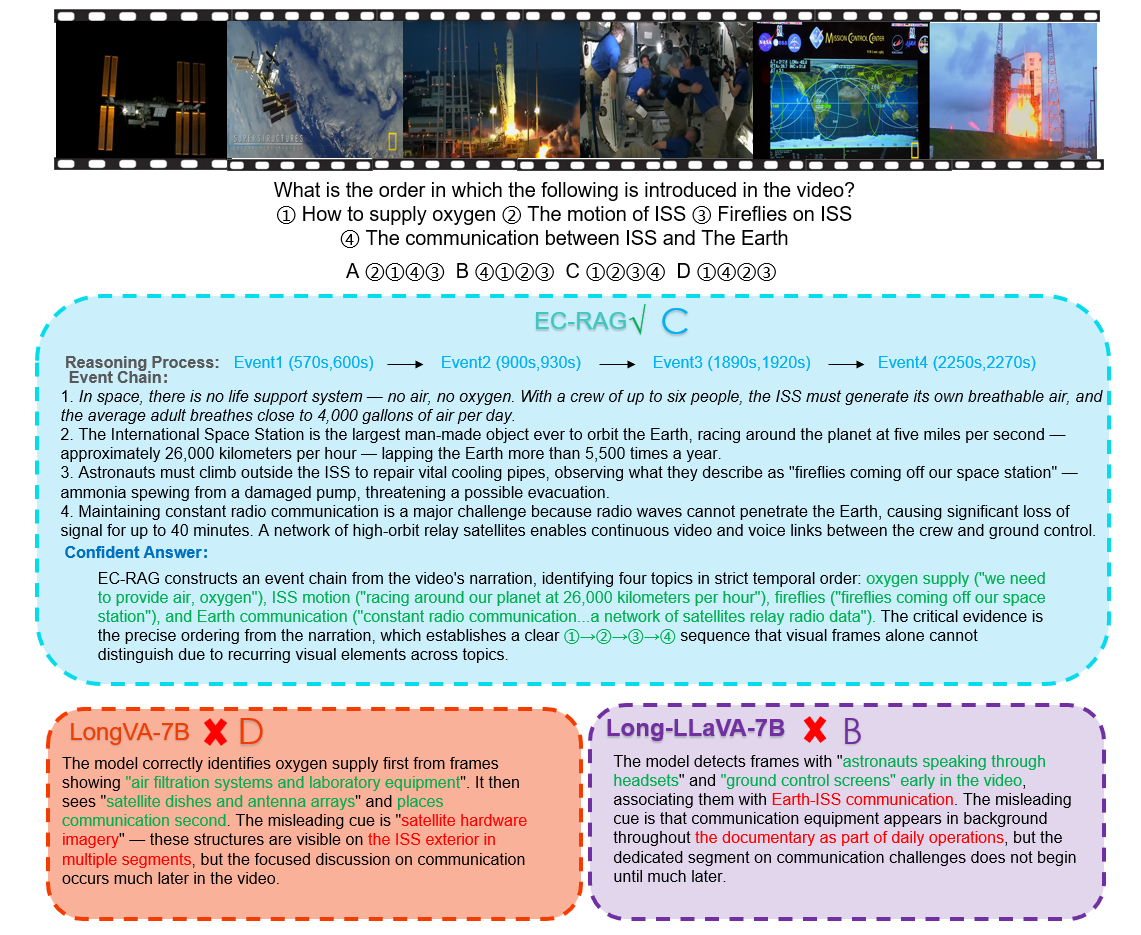}
  \caption{Qualitative example 2: a science documentary where the question asks about the introduction order of four topics. The baseline is misled by recurring visual elements across segments, while EC-RAG recovers the correct topic sequence from the structured event chain.}
  \Description{Qualitative comparison between baseline and EC-RAG on a science documentary topic ordering question.}
  \label{fig:qualitative_case2}
\end{figure*}

\section{More Qualitative Results}
\label{sec:appendix_qualitative}

We present additional qualitative results of LLaVA-Video-7B~\cite{zhang2024llava} with EC-RAG on two representative examples from Video-MME~\cite{fu2025video} in Figures~\ref{fig:qualitative_case1} and~\ref{fig:qualitative_case2}. Both cases involve temporal reasoning questions from long documentary videos, where the baseline model fails due to the inherent ambiguity of sparsely sampled frames. As illustrated, by constructing temporal event chains from multimodal evidence—particularly ASR transcripts that capture the chronological progression of topics and events—EC-RAG enables the model to accurately resolve temporal ordering and causal relationships that are indistinguishable from visual information alone. The results demonstrate that event-level retrieval effectively bridges the gap between sparse visual sampling and the dense temporal reasoning required for long video understanding.









\end{document}